\documentclass[10pt,conference]{IEEEtran}
\usepackage{dsfont}

\usepackage{color}
\usepackage{setspace}
\usepackage{clrscode}
\usepackage{amsthm}
\usepackage{amsmath,bm}
\usepackage{pict2e}
\usepackage{amsmath}
\usepackage{amssymb}
\usepackage{graphicx}
\usepackage{bbold}
\usepackage[ruled,vlined]{algorithm2e}
\usepackage{caption}
\usepackage{subcaption}
\usepackage{mathabx}
\usepackage{stfloats}
\usepackage{multirow} % This one is used for tables
\usepackage{float} % This one is used for tables
\usepackage[ps2pdf,
bookmarks=false,
bookmarksnumbered=false, % true means bookmarks in
bookmarksopen=false, % true means only level 1
colorlinks=false]{}

\DeclareMathOperator*{\argmax}{argmax}
\DeclareMathOperator*{\argmin}{argmin}

\newcommand{\mV}{\mathcal{V}}
\newcommand{\mL}{\mathcal{L}}
\newcommand{\mT}{\mathcal{T}}
\newcommand{\mS}{\mathcal{S}}

\newcommand{\svec}{\textbf{s}}

\newcommand{\avec}{\textbf{a}}
\newcommand{\vvec}{\textbf{v}}
\newcommand{\vhvec}{\hat{\textbf{v}}}
\newcommand{\shvec}{\hat{\textbf{s}}}
\newcommand{\stvec}{\widetilde{\textbf{s}}}
\newcommand{\pvec}{\textbf{p}}

\newcommand{\R}{\mathbb{R}}

\newcommand{\Pb}{\text{P}}

\newcommand{\dt}{\Delta t}
\newcommand{\xibm}{\bm{\xi}}

\allowdisplaybreaks[2]

\begin{document}
%	\title{Learning-Based Path Planning for Multi-UAVs in the Presence of Jammers}
	\title{Jamming-Resilient Path Planning for Multiple UAVs via Deep Reinforcement Learning}
	
	\author{Xueyuan Wang\IEEEauthorrefmark{1}, M. Cenk Gursoy\IEEEauthorrefmark{1}, Tugba Erpek\IEEEauthorrefmark{2} and Yalin E. Sagduyu\IEEEauthorrefmark{2}
		\\ \IEEEauthorrefmark{1}Department of Electrical Engineering and Computer Science,
		Syracuse University, Syracuse, NY 13244
		\\		\IEEEauthorrefmark{2}  Intelligent Automation, Inc., Rockville, MD 20855
		\\Email: xwang173@syr.edu, mcgursoy@syr.edu, terpek@i-a-i.com, ysagduyu@i-a-i.com}

%	\author{\IEEEauthorblockN{Xueyuan Wang and M. Cenk Gursoy}
%		\thanks{The authors are with the Department of Electrical
%			Engineering and Computer Science, Syracuse University, Syracuse, NY, 13244
%			(e-mail: xwang173@syr.edu,  mcgursoy@syr.edu).}
%		\thanks{Manuscript received May 27, 2019; revised September 1, 2019.}}
	
	\maketitle
	
	\begin{abstract}\let\thefootnote\relax\footnotetext{This work was supported in part by the  National Science Foundation under grant CCF-1618615.}
		\let\thefootnote\relax\footnotetext{The material in this paper will be presented at the IEEE International Conference on Communications (ICC) 2021.}
		Unmanned aerial vehicles (UAVs) are expected to be an integral part of wireless networks. In this paper, we aim to find collision-free paths for multiple cellular-connected UAVs, while satisfying requirements of connectivity with ground base stations (GBSs) in the presence of a dynamic jammer. We first formulate the problem as a sequential  decision making problem in discrete domain, with  connectivity, collision avoidance, and kinematic constraints. We, then, propose an offline temporal difference (TD) learning algorithm with online signal-to-interference-plus-noise ratio (SINR) mapping  to solve the problem. More specifically, a value network is constructed and trained offline by TD method  to encode the interactions among the UAVs and between the UAVs and the environment; and an online SINR mapping deep neural network (DNN) is designed and trained by supervised learning, to encode the influence and changes due to the jammer.
		Numerical results show that, without any information on the jammer, the proposed algorithm can achieve performance levels close to that of the ideal scenario with the perfect SINR-map. Real-time navigation for multi-UAVs can be efficiently performed with high success rates, and collisions are avoided.
	\end{abstract}


	\thispagestyle{empty}

	\section{Introduction}

	Unmanned aerial vehicles (UAVs), also commonly known as drones, have found numerous applications and are expected to be utilized extensively in different use cases in the next decade \cite{UAV_survey_YZeng}. UAVs are also  predicted to be a critical part of future wireless communication networks. For instance, in order to take advantage of flexible deployment opportunities and high possibility of line-of-sight (LoS) connections	with ground user equipments (UEs), UAVs can be deployed as aerial base stations (BSs) to support wireless connectivity	and improve the performance of cellular networks \cite{liu2018performance}, leading to a UAV-assisted cellular network architecture. On the other hand, UAVs in certain applications will 	be regarded as aerial UEs that need to be supported by the ground communication	infrastructure, leading to a cellular-connected UAV network architecture \cite{UAV_cellular_MAzari}.

	However, the UAV-enabled wireless communication systems can be easily vulnerable to jamming and eavesdropping attacks due to the broadcast nature of wireless transmissions \cite{jamming_wang2018survey}. In particular, jamming is a malicious attack whose objective is to disrupt the communication in the victim network by intentionally causing interference  at the receiver side \cite{jamming_bhattacharya2010game}.
	Therefore, once attacked, the quality of communication will decline, leading even to link loss and mission interruptions.
	A series of strategies to resist  jamming attacks have been proposed in wireless networks, which are generally divided into two categories: 1) adapting to the jamming signal; 2) retreating away from or avoiding the jammer \cite{jamming_duan2018anti}.
%	As for retreat,	\cite{jamming_xu2004channel}  proposed two evasion approaches: channel	surfing and spatial retreats. Channel surfing involved legitimate	wireless nodes changing communication channel, and spatial	retreats belongs to spatial evasion where the jammed node	selects randomly the direction of movement until escaping
%	from the jammed area. As for competing with jammer, \cite{jamming_lv2017anti} presented that legitimate nodes adjust transmit power	level to make receivers get higher signal-to-interference-plusnoise	ratio (SINR).

	In the first category, physical-layer security has emerged as a promising approach to secure UAV communications against jamming attacks.
	%, such as  beamforming that can be used to attenuate transmissions from certain angles \cite{jamming_abdalla2020uav}, power allocation strategies.
	For instance, authors in \cite{jamming_xiao2017user}  proposed deep Q-learning  based UAV power allocation strategies to improve the static UAV-enabled communications against smart jamming attacks.
	In \cite{jamming_wang2020energy},  a reinforcement learning based power control algorithm was proposed to improve the performance of the multi-UAV relay communication systems in the presence of a random jammer.
	The authors in \cite{jamming_zhou2019uav}  introduced  a UAV as a friendly jammer in the UAV-enabled network to work against multiple eavesdroppers, aiming to maximize the minimum average secrecy rate over all information receivers.

	%\cite{jamming_lu2020uav} considered to employ UAVs as mobile relays to facilitate secure or reliable wireless communications. 	
	%\cite{jamming_li2018uav, jamming_zhou2017secrecy, jamming_zhong2018secure} proposed to use UAVs as friendly jammers to secure the ground wireless communication.

	Owing to the mobility and flexibility of UAVs, it is also feasible to use the avoidance strategy against jamming attacks.
	Furthermore, spatial evasion based methods do not impose high requirements on communications devices \cite{jamming_duan2018anti}.
	By taking advantage of the  mobility of UAVs, jamming-resistant trajectory designs in UAV-enabled communication systems have been studied in the literature.
	For instance, authors in \cite{jamming_gao2020robust} investigated the maximization of the uplink data throughput of UAV-enabled communication in presence of a potential jammer. An alternating algorithm that leverages the block coordinate descent method, successive convex approximation, and S-procedure, was proposed to optimize the UAV trajectory.  A similar problem was studied in \cite{jamming_wu2019robust}, where the goal is to improve the minimum uplink data throughput for multi-UAV enabled communication in the presence of jammers with imperfect location information.
	In \cite{jamming_duo2020anti}, the authors aimed to maximize the minimum (average) expected data collection rate from ground sensors in the presence of a malicious ground jammer, by jointly optimizing the ground sensor transmission schedule and UAV horizontal and vertical trajectories over a finite flight duration. %An iterative algorithm was proposed to solve the problem. 	
	In \cite{jamming_wang2018trajectory}, the authors had the objective to maximize the sum throughput received by the UAV in the presence of jammer signals, by designing the UAV deployment and trajectory planning in three-dimensional (3D) space.
%	In \cite{jamming_wang2020energy}, by optimizing the UAV trajectory, the authors aimed to maximize the energy efficiency of the UAV when communicating with ground nodes in the presence of multiple jammers.
	\cite{jamming_lin2019reinforcement} proposed a deep Q-learning based UAV trajectory and power control scheme against smart jamming attacks on transmissions with ground nodes, given the predefined UAV sensing waypoints.
	The prior work in this area has mostly concentrated on either single-UAV scenarios or multiple UAVs operating as aerial BSs, and hence has not addressed cellular-connected UAVs, connectivity constraints and collisions avoidance requirements especially with a jammer present, using a learning framework.
	
	Motivated by these facts, extending our recent work in \cite{traj_ICC_XWang}, we propose a deep reinforcement learning (RL) based algorithm to perform path planning for multiple cellular-connected UAVs with wireless connectivity and collision avoidance constraints in the presence of dynamic jamming attacks. We note that as key novelties over \cite{traj_ICC_XWang}, we consider jamming attacks in this paper and build a novel online SINR mapping algorithm in addition to modifying the value network of the RL agent to address the jamming interference. We further note that the proposed algorithm does not require any prior information on the jammer.

	\section{System Model and Problem Formulation}
	
	\begin{figure}
		\centering		
		\includegraphics[width=0.35\textwidth]{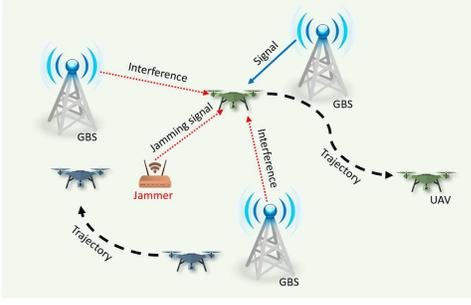}			
		\caption{\small An illustration of the cellular networks with multiple UAVs, multiple GBSs, and a jammer.\normalsize}
		\label{Fig:network}
	\end{figure}
	\subsection{System Model}
	A multi-UAV multi-ground base station (GBS) cellular network is considered, in which $K$ GBSs provide wireless connectivity to the UAVs. 		
	There is also a jammer in the network that transmits jamming signals to interfere the links between the UAVs and their serving GBSs, and hence disrupts the communication. The location and the transmit power of the jammer can vary over time.
	
	Multiple UAVs, with different missions, need to fly from starting points to their destinations. Each UAV's state is composed of an observable information vector and an unobservable (hidden) information vector, $\svec = [\svec^o, \svec^h]$, where the observable state can be observed by other UAVs, while the unobservable state can not. In the global frame, observable state includes the UAV's position $\pvec = [p_x,p_y,H_V]$, velocity $\vvec = [v_x,v_y]$, and radius $r$, i.e., $\svec^o = [\pvec, \vvec, r] \in \R^6$. The unobservable state consists of the destination $\pvec^D$, maximum speed $v_{\max}$, and orientation $\phi$, i.e., $\svec^h =[\pvec^D, v_{\max},\phi] \in \R^5$. It is worth noting that the UAVs do not communicate with other UAVs. Each UAV is assumed to be associated with the GBS providing the largest received power. Since the UAV is moving, the associated GBS changes over time.    An illustration of the cellular networks with multiple UAVs, multiple GBSs, and a jammer is provided in Fig. \ref{Fig:network}.
	
	The UAVs receive desired signal from the serving GBS,  interference from other GBSs and jamming signal from the jammer. Thus, the experienced SINR at a UAV can be expressed as 	
	\begin{align}
	\label{Eq:SINR}
	\mS_{r_{k}} \triangleq \frac{P_k G_{B}(d_{k}) G_{V}(d_{k}) L^{-1}(d_{k}) }{\mathcal{N}_s  + I_J(d_J)+ \sum_{k' \neq k}^{K}P_{k'} G_{B}(d_{{k'}}) G_{V}(d_{{k'}}) L^{-1}(d_{{k'}}) }
	\end{align}
	where $\mathcal{N}_s $ is the noise power, $P_k$ is the transmit power of the $k^{th}$ GBS, $d_k $ is the horizontal distance between the UAV and the $k^{th}$ GBS.
	$G_B$ is the 3D antenna gain at the GBSs, and can be formulated as
	\begin{align}
	G_B(d)
	= 10 ^{- \min\left(-1.2 \left(\frac{\arctan\left(\frac{H_B -H_V}{d}\right)-\theta^{tilt}}{\theta^{3dB}}\right)^2, \frac{G_m}{10}\right)} 	.
	\end{align}
	$G_V$ is the 3D antenna gain at the UAVs, and is expressed as
	\begin{align}
	G_V (d) = \sin(\theta) = \frac{H_V- H_B}{\sqrt{d^2 + (H_V -H_B)^2}}.
	\end{align}
	$\theta$ is elevation angle between the UAV and GBS, $H_V$ is the UAV altitude, and $H_B$ is the height of the GBS.
	$L$ is the path loss
	\begin{align}
	L(d) =  \left(d^2 + (H_B-H_V)^2 \right) ^{\alpha/2}
	\end{align}
	where $\alpha$ is the path loss exponent.
	In addition, $I_J$ is the interference from the jammer, which can be expressed as
	\begin{align}
	I_J(d_J) =  P_J\left(d_J^2 + (H_J-H_V)^2 \right) ^{\alpha/2}\frac{ (H_V- H_J) }{\sqrt{d_J^2 + (H_V -H_J)^2}}
	\end{align}
	where $P_J$ and  $H_J$  are the transmit power and height of the jammer, respectively, and $d_J$ is the horizontal distance between the UAV and the jammer. If the SINR experienced at the UAV is smaller than a threshold $\mT_s$, the UAV is regarded as disconnected from the network.

	\subsection{Problem Formulation}
	
	The goal of this work is to find policies to determine the trajectories for  UAVs such that the mission completion time is minimized and the constraints are satisfied. The main constraints considered in this paper include the following: wireless connectivity constraint, collision avoidance constraint, and kinematic constraints. More specifically, wireless connectivity constraint imposes that to support the command and control and also data flows, the UAVs have to maintain a reliable communication link to the GBSs. To achieve this goal, the maximum contiguous time period that the UAVs are disconnected from the cellular network should not be longer than a threshold $\mT_t$.  The collision avoidance constraint introduces the requirement that during flight, the distance between two UAVs should not be smaller than the sum of their radii all the time. The kinematic constraints restrict the speed and the rotation angle of the UAVs, i.e., the rotation angle of a UAV in unit time period and its speed are limited to a certain range.

	 We assume that in the network there are in total $J$ UAVs with wireless connectivity constraints.  We use a common assumption that the UAVs in the network follow the same policy to find their trajectories \cite{van2011reciprocal,chen2016decentralized,CA_HKretzschmar}. Without loss of generality, we choose the $i^{th}$ UAV as the typical one, and formulate the problem in the discrete time domain as follows:
	\begin{align}
	(\Pb 1):\argmin_{ \{\pvec_{i,t},  \forall t\}}  & \qquad T_i \notag  \\
	s.t.         \quad
	& \mS_{r_{i,t}} \geq \mT_s, \text{ if } t \mid n_t \tag{P1.a}\\
	& ||\pvec_{i,t} - \pvec_{j,t}  ||_2 > r_i+r_j, \forall j \neq i, \forall t \tag{P1.b}\\
	& \pvec_{i,0} = \pvec_i^S, \pvec_{i,T_i} = \pvec_i^{D}, \forall i \tag{P1.c}\\
	& v_{s_{i,t}} \leq v_{\max_i}, \forall t \tag{P1.d}\\
	& |\phi_{i,t}- \phi_{i,t-1}| \leq \Delta t \cdot \mT_r, \forall t \tag{P1.e}
	\end{align}
	where the integer-valued discrete time index $t$ indicates time increments by $\Delta t$ (a constant time period), and $\pvec_{i,t}$ is the position of the $i^{th}$ UAV at time step $t$. (P1.a) is due to the connectivity constraint, and $t \mid n_t$ signifies that $t$ is divisible by  $n_t = \mT_t/\Delta t$. In (P1.a), we have converted the maximum continuous disconnection time constraint into  discrete form by checking the SINR every $n_t$ time steps. (P1.b) is the collision avoidance constraint, which is important in multi-UAV scenarios.  (P1.c) is the constraint indicating the start and destination locations. (P1.d) and (P1.e) arise from the kinematic constraints.

	\section{Proposed Algorithm}	
	The proposed optimization problem is difficult to solve due to the non-convex constraints, lack of knowledge on the jammer, and the interactions among multiple UAVs. To overcome this difficulty, we can cast the problem into a sequential decision making problem that can be solved by reinforcement learning (RL).
%\subsection{Reinforcement Learning}
%   RL is a class of machine learning methods for solving sequential decision making problems with unknown state-transition dynamics \cite{chen2016decentralized}  \cite{everett2020collision}.
	Typically, a sequential decision making problem can be formulated as a Markov decision process (MDP) \cite{RL_MIT}, which is described by the tuple $\langle S,A,P,R,\gamma \rangle $, where $S$ is the state space, $A$ is action space, $P$ is the state-transition model, $R$ is the reward function, and $\gamma$ is the discount factor.	
	The essential task of many RL algorithms is to obtain the optimal value function, which can be written as
	\begin{align}
		V^* (\svec^{jn}_t) = \sum^T_{t'=t} \gamma^{t'-t} R_{t'}(\svec^{jn}_{t'}, \pi^*(\svec^{jn}_{t'})).
	\end{align}
%	which satisfy the celebrated Bellman optimality equation
%	\begin{align}
%	&V^* (\svec^{jn}_t) = \notag \\
%	& \max_{\vvec_t} R(\svec_{t}^{jn},\vvec_t)
%	+ \gamma  \int_{\svec_{t+1}^{jn}} P(\svec_{t+1}^{jn}|\svec_{t}^{jn},\vvec_t)  V^*(\svec_{t+1}^{jn}) \text{d}\svec_{t+1}^{jn}
%	\end{align}
	The optimal policy is the one that maximizes the expected return, and can be expressed as
	\begin{align}\label{Eq:optimal_policy}
	\pi^*(\svec_{t}^{jn}) &= \argmax_{\vvec_t} R(\svec_{t}^{jn},\vvec_t) \notag \\
	&+ \gamma  \int_{\svec_{t+1}^{jn}} P(\svec_{t+1}^{jn}|\svec_{t}^{jn},\vvec_t)  V^*(\svec_{t+1}^{jn}) \text{d}\svec_{t+1}^{jn},
	\end{align}	
	where $\svec_{t}^{jn}$ is the joint state at time $t$, $\vvec_t$ is the action, and $R(\svec_{t}^{jn},\vvec_t)$ is the reward received at time $t$, $P(\svec_{t+1}^{jn}|\svec_{t}^{jn},\vvec_t)$ is the transition probability from time $t$ to time $t+1$.
		
	When estimating the high-dimensional, continuous value function, it is common to approximate it using a deep neural network (DNN) parameterized by weights and biases, $\xibm $.  %For notational simplicity, we drop the DNN parameters from the value function notation,  i.e., $\mV(\svec;\xibm ) = \mV(\svec)$. And
	Given the joint state of an agent $\svec$, which is also the input of the DNN, the output of the value network is denoted by $\mV(\svec;\xibm)$.
	
%\subsection{Proposed Algorithm}
	Model-free RL requires no prior knowledge about the environment. This usually leads to slow learning process and requires a large number of agent-environment interactions, which is typically costly or even risky to obtain \cite{uavtraj_YZeng}. In addition, the existence of the dynamic jammer makes the environment vary frequently, and thus even harder to encode and learn.  On the other hand, online learning (interactions with the real environment) for multi-UAV scenario is costly due to high collision risk. Therefore, we propose a offline temporal-difference (TD) algorithm with online SINR mapping  for multi-UAV path  planning with jamming resiliency.
	The  proposed algorithm consists of two modules: 1) offline value network training by TD method with standard experience replay; 2) online SINR mapping by supervised learning; and both modules will be introduced in detail in the following subsections.
%	\begin{itemize}
%		\item Off-line value network training: in the simulator, the jammer are assumed to be located in some positions. A value network is trained assume perfect knowledge of the radio map.
%		\item A SINR-prediction NN is trained on-line in a cloud. It is assumed the cloud has access to  the (location, SINR) pairs from all UAVs flying in the environment. After a certain period, the SINR-prediction NN is updated according to the pairs measured by the UAVs.
%	\end{itemize}
%		in the jamming Actually, each real experience obtained from the UAVs and cellular network interaction not only can be used to get reward and refine the value network, but also can be used for model learning in order to predict the agent's SINR experienced at certain positions.

\subsection{Offline Value Network Training}
	This offline learning module can be implemented on a simulator, therefore reducing the collision risk.  The environment can be modeled close to the reality, containing the following information:  GBSs which are distributed according to the real deployment; the channel  modeled according to real ray-tracing data; a dynamic jammer, which changes its location and transmit power periodically.
	
	TD learning is used to train the value network, and the $\langle S,A,P,R,\gamma \rangle $ formulation for the proposed problem is provided as follows:
	\subsubsection{State}
	In multi-UAV cellular networks, the UAVs are able to observe the following information from the environment: 1) its own information vector $\svec_{i,t}$ (for the $i^{th}$ UAV at time step $t$); 2) the observable state of the nearest $J_n < J$ UAVs $\svec_{i,t}^{jno} = [ \svec_{j,t}^o: j\in \{1,2,...,J_n \} ]$; 3) the  experienced SINR $\mS_{r_{i,t}}$.	
	Since the policy should not be influenced by the choice of the coordinates, we choose an agent-centric coordinate plane, where the UAV's location is the origin. Then, we need to  change the positions from the global frame to the chosen coordinates. In addition, the observed information can be parameterized to provide more information.  Hence, the observations are transformed into
	\begin{align}
	&\stvec_i =  [v_{x_i}, v_{y_i}, \tilde{p}_{d_{x_i}}, \tilde{p}_{d_{y_i}},  d_{d_i},  a_{d_i}, r_i, v_{\max_i}, \phi_i] \notag \\
	& \stvec^{jno}_i =   [[\tilde{p}_{x_j}, \tilde{p}_{y_j}, v_{x_j}, v_{y_j},  d_{j}, a_{j}]: j \in \{ 1,2,...,J_n  \} ]  \notag \\
	& \tilde{\mS}_{r_{i,t}} = L_{w_{i,t}} \notag
	\end{align}
	where $\tilde{p}$ denotes $p$ in the new coordinates. $d_{d_i}$ and $a_{d_i}$ are the distance and azimuth angle from the typical UAV to its destination. $d_j$ and $a_j$ are the distance and azimuth angle to the $j^{th}$ other UAV.  $L_{w_{i,t}}$ is the quantized SINR level.
	All the information observed by the agent constitutes its joint state
	\begin{align}
	\label{Eq:state}
	\svec_{i,t}^{jn} = [\stvec_{i,t}, \stvec_{i,t}^{jno}, \tilde{\mS}_{r_{i,t}}], \forall t.
	\end{align}

	\subsubsection{Action}
	Based on the agent's current speed, orientation $[v_{s,i,t},\phi_{s,i,t}]$ and the kinematic constraints, permissible actions $\vvec=[v_s, \phi]$ are sampled to build the action space $A_{i,t}$.
	
	\subsubsection{Reward}
	To encourage being connected to the cellular network, a  reward $R_s$ is designed as a step function that can be expressed as follows:
	\begin{align}
	&R_{s_{i,t}}(\svec_{i,t}^{jn}, \vvec_{i,t}) = \notag \\
	&\begin{cases}
	-0.5,  &\text{if }  t\mid n_t  \text{ and }  \mT_s \leq \mS_{r_{i,t+1}}< \mT_s + 0.1 \\\
	-1,   &\text{if }  t \mid n_t \text{ and }  \mS_{r_{i,t+1}} < \mT_s \\
	0, & \text{otherwise}.
	\end{cases}
	\end{align}
	To encourage not getting close to or not colliding with other agents, reward $R_c$ is designed as a function of the minimum distance to other UAVs. Specifically, $R_c$ is formulated as
	\begin{align}
	&R_{c_{i,t}}(\svec_{i,t}^{jn}, \vvec_{i,t}) = \notag \\
	&\begin{cases}
	-1,                         & \text{if } d_{t_{\min}}\leq r_i+r_j,\\
	- \left(1-  \frac{d_{t_{\min}}-r_i-r_j}{0.2} \right), & r_i+r_j<d_{t_{\min}} \leq 0.2+r_i + r_j, \\	
	0, & \text{otherwise},
	\end{cases}
	\end{align}
	where  $d_{t_{\min}}$ is the minimum distance to other agents within the next time  duration.
	To award the arrival to the destination, there is a reward $R_{d_{i,t}}(\svec_{i,t}^{jn}, \vvec_{i,t}) = 2$ if $ \pvec_{i,t+1} = \pvec_i^D$.
%	To award the arrival to the destination, there is also the following additional reward term:
%	\begin{align}
%	R_{d_{i,t}}(\svec_{i,t}^{jn}, \vvec_{i,t}) =
%	\begin{cases}
%	2,                             & \text{if } \pvec_i = \pvec_i^D, \\
%	0, & \text{otherwise},
%	\end{cases}
%	\end{align}
	To encourage fast arrival to the destination, a constant movement penalty, $R_t$,  is given as well. Finally,  the overall reward function can be expressed as the sum
	\begin{align}
	&R_{i,t}(\svec_{i,t}^{jn}, \vvec_{i,t}) = \notag \\
	&R_{c_{i,t}}(\svec_{i,t}^{jn}, \vvec_{i,t}) + R_{s_{i,t}}(\svec_{i,t}^{jn}, \vvec_{i,t}) + R_{d_{i,t}}(\svec_{i,t}^{jn}, \vvec_{i,t}) + R_{t}.
	\end{align}
	
%	\subsubsection{Probabilistic State Transition Model}
%	Since the probabilistic state transition model is determined by the UAV's kinematics, others agents' hidden states and their choices of action, which are not known, the state transition model is not known. We adopt a one-step lookahead procedure as in \cite{rvo_JBerg,van2011reciprocal,chen2016decentralized,hrvo_JSnape}, i.e. it is assumed  that the other agent would be traveling at a filtered velocity for a short duration $\dt$.
	
	 A number of  trajectories are generated by optimal reciprocal collision avoidance (ORCA) \cite{van2011reciprocal}, in order to obtain a set of state-value pairs $D$ to initialize the value network $\xibm$. Overall, the offline value network training is summarized in Algorithm \ref{Algm:main_algm}.  In line 16 of Algorithm \ref{Algm:main_algm}, the next state $\shvec_{i,t+1}^{jno}$ is obtained by one-step lookahead procedure as in \cite{van2011reciprocal,chen2016decentralized,rvo_JBerg,hrvo_JSnape}, i.e., it is assumed  that the other agents would be traveling at a filtered velocity for a short duration $\dt$.
	\begin{algorithm}
		\caption{Offline Value Network Learning}
		\label{Algm:main_algm}
		\LinesNumbered
		\KwIn{State-value pairs $D$}
		%	\KwOut{$\mV, \mL$}
%		Initialize state-value pairs  $D$\\
%		Initialize location-SINR pairs $D_w$\\
		Initialize value network $\xibm$ with $D$\\
%		Initialize value network $\xibm$\\
%		Initialize SINR-prediction network $\xibm_w$	\\		
		%			\ENSURE trajectory $\svec_{0:T}$
		\For{episode = 0: total episode}{			
%				Initialize  $\svec_{i,0} \forall i$ \\
				Reset environment and jammer \\
				\While{not all reached destinations}{
					\For{each agent $i$}{
						\If{not reached destination}{
							%Observe the environment, and update $\svec_{i,t}^{jn}$ \\
							$\svec_{i,t}^{jn} \leftarrow \text{observeEnvironment}()$ \\
							$A_{i,t} \leftarrow \text{sampleActionSpace}()$ \\
							%Sample $c$ from Uniform (0,1) \\
							$c \leftarrow \text{randomSample(Uniform (0,1))}$\\
							\eIf{ $c\leq \epsilon$ }{
								$\vvec_{i,t} \leftarrow \text{randomSample} (A_{i,t})$ }{
								$\vhvec_{i,t}^{jno} \leftarrow \text{filter}(\vvec_{0:t-1}^{jn})$\\
								$\shvec_{i,t+1}^{jno} \leftarrow \text{propagate}(\svec_{i,t}^{jno},
								\vhvec_{i,t}^{jno})$ \\
								\For{every $\avec$ in $A_{i,t}$}{
									$\shvec_{i,t+1} \leftarrow \text{propagate}(\svec_{i,t},
									\avec)$ \\
									$\hat{L}_{w_{i,t+1}} \leftarrow \text{radioMap}()$\\
									$R_{i,t} \leftarrow \text{estReward}(\shvec_{i,t+1}^{jn}, \hat{L}_{w_{i,t+1}})$\\
									$V_{p} = R_{i,t} + \gamma \mV (\shvec_{i,t+1}^{jn};\xibm)$
								} % endfor
								$\vvec_{i,t}  \leftarrow \argmax_{\avec\in A_{i,t}} V_{p}$
							} % endif
							$R_{i,t}, \svec_{i,t+1}, \mS_{r_{i,t+1}} \leftarrow \text{executeAction}(\vvec_{i,t} )$						
				}}}	
				\For{each agent $i$}{		
					$V_{i,0:T_i} \leftarrow \text{updateValue}(\svec^{jn}_{i,0:T_i},R_{i,0:T_i},\xibm) $		\\
%					$L_{w_{i,0:T_i}}  \leftarrow \text{getSINRlevel}(\mS_{r_{i,0:T_i}})$ \\
					Update state-value pairs $D$ with $\langle\svec_{i,0:T_i}^{jn}, V_{i,0:T_i} \rangle$\\		
%					Update location-SINR pairs $D_w$ with $\langle[\pvec_{i,0:T_i}, \svec_B],L_{w_{i,0:T_i}}  \rangle$
				}			
			Sample random minibatch from $D$, and update value network $\xibm$ by gradient descent.\\
		}
		\Return $\xibm$
	\end{algorithm}

\subsection{Online SINR Mapping}
   Along the path to destination, UAVs interact with the cellular network, measure the raw signal from GBSs, and obtain the instantaneous SINR. The measurement can be obtained by leveraging the existing soft handover mechanisms with continuous reference signal received power (RSRP) and reference signal received quality (RSRQ) \cite{uavtraj_YZeng}. The empirical instantaneous SINR can be processed to obtain the SINR in (\ref{Eq:SINR}). UAVs with sensors can also observe the positions of the nearby GBSs $\svec^{jn}_{B} = [\pvec_{B_k}: k \in \{1,...,K_n\} ]$. Therefore, the UAVs with sensors are able to obtain measurements  $\{\langle \svec_{B}^{jn}, \mS_r(\svec_{B}^{jn}) \rangle \}$  along their paths.

   It is assumed that there is a cloud computing center to where all UAVs in the area can upload their measurements, and the cloud will keep updating the recent measurements in its memory $D_w$.   A DNN, denoted by $\xibm_w$, can be designed to map the position information of nearby GBSs into the SINR level by supervised learning using the recent measurements in memory. The DNN is updated periodically based on how frequently the jammer varies.
   \subsubsection{Input}
   The input of the DNN is the parameterized position information vector of the nearby GBSs.   By changing the positions into the agent-centric coordinates and processing relative locations into distance and angles, the vector can be transformed into
   \begin{align}
   %	\svec^{jn}_{B_i} = [\pvec_i, \svec^o_{B}]  \rightarrow
   \stvec^{jn}_{B_i} =  [[\tilde{p}_{k_{x_k}},\tilde{p}_{k_{y_k}}, d_{B_k}, \phi_{B_k}, \theta_{B_k}]: k \in \{1,...,K_n\}   ]
   \end{align}
   where $d_{B_k}$, $\phi_{B_k}$, $\theta_{B_k}$ are the distance, elevation angle, azimuth angle from the UAV to the $k^{th}$ nearby GBS.
   \subsubsection{Label}
   The label of the supervised DNN is the quantized SINR level $L_w(\svec_{B}^{jn})$.
   \subsubsection{Detection}
    A simple jammer change detection method can be adopted. More specifically, the cloud can check the accuracy of the DNN periodically on the newly uploaded measurements. The accuracy dropping significantly indicates the changes in the jammer (either its location or transmit power). Then, the DNN should be updated. From numerical results, this jammer change detection method takes less than 1s.
   \subsubsection{Training}

   If the parameters of the DNN $\xibm_w$ need to be updated,  training can be done by stochastic gradient descent (back-propagation) on mini-batches randomly sampled from $D_w$ for a fixed number of episodes.

\subsection{Real-Time Navigation}
The UAVs can perform real-time navigation with the offline trained value network and online SINR mapping DNN downloaded from the computing cloud. 	Lines 15- 20 in Algorithm \ref{Algm:main_algm} can be used to choose actions in each time step in real-time navigation, except that in line 19, $\shvec^{jn}_{B_{i,t+1}}$ should be estimated first, and then $\hat{L}_{w_{t+1}}$ is given by the SINR mapping DNN, i.e.,  $\hat{L}_{w_{i,t+1}} = \mL_w(\shvec^{jn}_{B_{i,t+1}}; \xibm_w)$.

%	\begin{algorithm}
%		\caption{Real-Time Navigation}
%		\label{Algm:execution}
%		\LinesNumbered
%		\KwIn{$\xibm$}
%		Update SINR mapping DNN $\xibm_w$ \\
%		Initialize $\svec_{0}$ \\
%		\While{not reached destination}{
%			%Observe the environment, and update $\svec_{i,t}^{jn}$ \\
%			$\svec_{t}^{jn} \leftarrow \text{observeEnvironment}()$ \\
%			$A_{t} \leftarrow \text{sampleActionSpace}()$ \\					
%			$\vhvec_{t}^{jn} \leftarrow \text{filter}(\vvec_{0:t-1}^{jn})$\\
%			$\shvec_{t+1}^{jno} \leftarrow \text{propagate}(\svec_{t}^{jno},
%			\vhvec_t^{jn})$ \\
%			\For{every $\avec$ in $A_{t}$}{
%				$\shvec_{t+1} \leftarrow \text{propagate}(\svec_{t},
%				\avec)$ \\
%				$\hat{L}_{w_{t+1}} = \mL_w([\phvec_{t+1}, \svec_B])$\\
%				$R_{t} \leftarrow   \text{getReward}(\shvec_{t+1}, \shvec_{t+1}^{jno}, \hat{L}_{w_{t+1}})$\\
%				$V_{p} = R_{t} + \gamma \mV (\shvec_{t+1}^{jn})$
%			} % endfor
%			$\vvec_{t}  \leftarrow \argmax_{\avec\in A_{t}} V_{p}$ \\
%			$\svec_{t+1}\leftarrow \text{executeAction}(\vvec_{t} )$											
%		}
%		\Return $\vvec_{0:T-1}$, $\svec_{0:T}$
%	\end{algorithm}

	\section{Numerical Results}
	In this section, we present the numerical results to evaluate the performance of the proposed algorithm.
	We consider an area  with 12 GBSs and 1 jammer. The GBSs have transmit power of $P_B=1$ W,  height of $H_B=32$ m, and the antenna patterns with $\theta^{tilt}=10^{\circ}$ and $\theta^{3dB} = 15^{\circ}$. The UAVs are assumed to fly at a fixed altitude of $H_V=50$ m. The noise power is $\mathcal{N}_s = 10^{-6}$, and the SINR threshold is $\mT_{s} = -3$ dB.  Each UAV, as an independent agent, is able to observe at most 4 other UAVs.
	
	The value network is designed as a DNN of size (64,32,16). The SINR mapping DNN is also  three-layered with size (32,16,8).  A standardization layer is utilized after the input layer of both networks.  ReLU activation function is used for the input layer and two hidden layer for both networks, while Tanh activation function is used for the output layer of the value network, and no activation function is used for the output layer of the SINR mapping DNN.
	Both networks use Adam optimizer,  batch size 200, and a regularization parameter 0.0001. The learning rates of the value network and SINR mapping DNN are 0.001 and  0.005, respectively. The exploration parameter $\epsilon$ linearly decays from $0.5$ to $0.1$. %The replay memory capacity is 30000 for the 2-agent scenario and 100000 for scenarios with more than two agents.
	
	\subsection{Training}
\begin{figure}
		\centering		
		\includegraphics[width=0.35\textwidth]{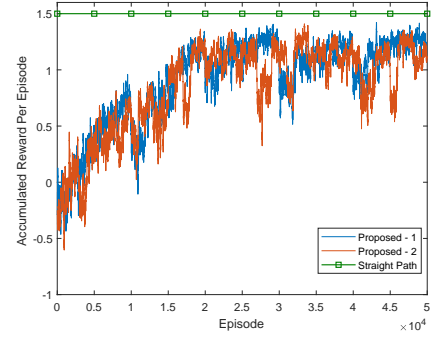}			
		\caption{\small Accumulated reward per episode for two distinct training cases and the straight path scenario, as functions of episode.\normalsize}
		\label{Fig:reward}
	\end{figure}

	In the offline value network training module, the jammer changes its location or transmit power periodically. Fig.  \ref{Fig:reward} shows the accumulated reward per episode during two distinct training. First, we can observe from the figure that the reward in two distinct training can converge to the same level. The relatively larger drops in the convergence phase (after 20000 episodes) are due to the significant changes in the jammer, and the value network can learn fast and recover the reward back to the converged level.  Secondly, we can observe that  our proposed algorithm can achieve comparable reward level to the upper bound. The upper bound in this figure (presented in green line with squared markers) is the accumulated reward if the UAVs fly straight to their destinations, ignoring the wireless connectivity (i.e. $R_{s_{i,t}}=0$) and collision avoidance constraints (i.e. $R_{c_{i,t}} = 0$). Note that when these constraints are taken into account,  the UAVs need to perform turns or stops. Therefore, due to the negative reward  $R_t$ given to each step and the exploration strategy (i.e. $\epsilon_{\min} =0.1$) , the reward achieved in training is smaller than the upper bound.
	
\begin{figure}
		\centering		
		\includegraphics[width=0.35\textwidth]{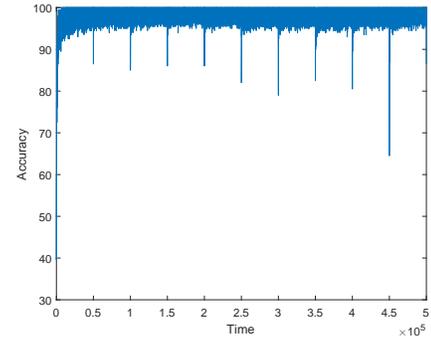}			
		\caption{\small Accuracy of the SINR mapping DNN as a function of training time.\normalsize}
		\label{Fig:acc}
	\end{figure}

	In the online SINR mapping module, the DNN is updated periodically using the latest uploaded SINR measurements. Fig. \ref{Fig:acc} presents the accuracy during the online training period. It can be observed that the DNN can adapt fast to the new location/transmit power of the jammer with very high SINR mapping accuracy.

	\subsection{Real-Time Navigation}
\begin{figure*}
		\centering
		\begin{minipage}{0.24\textwidth}
			\centering
			\includegraphics[width=1\textwidth]{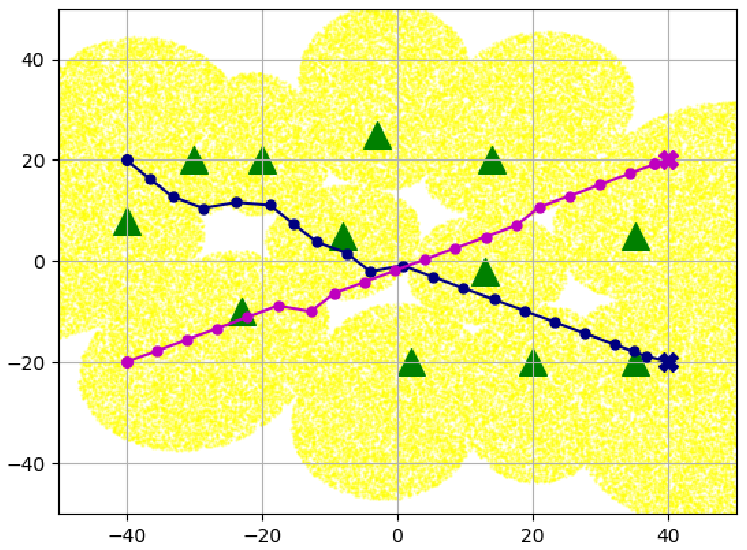}
			\subcaption{\scriptsize No jammer.   }
		\end{minipage}
		\begin{minipage}{0.24\textwidth}
			\centering
			\includegraphics[width=1\textwidth]{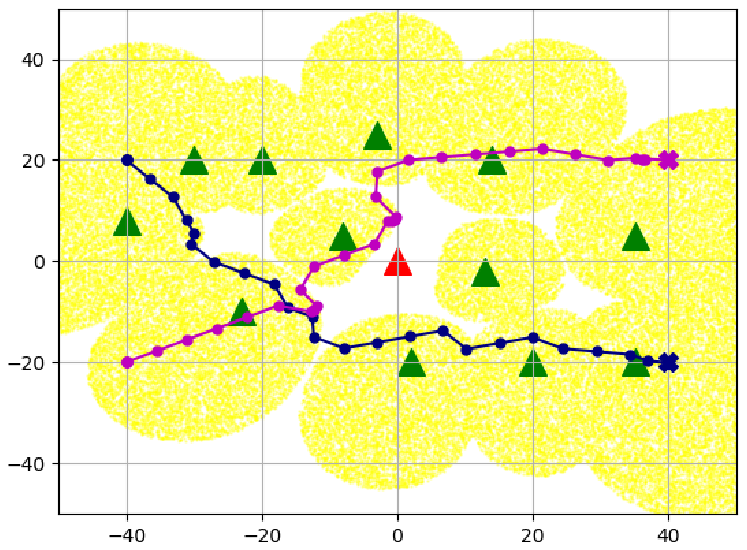}
			\subcaption{\scriptsize Jammer at (0,0) with $P_J=1$W.}
		\end{minipage}
		\begin{minipage}{0.24\textwidth}
			\centering
			\includegraphics[width=1\textwidth]{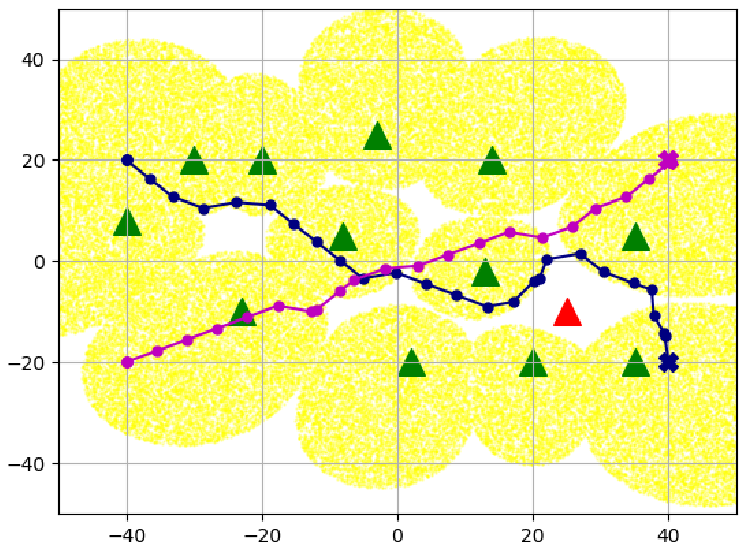}
			\subcaption{\scriptsize Jammer at (25,-10) with $P_J=1$W.}
		\end{minipage}
		\begin{minipage}{0.24\textwidth}
			\centering
			\includegraphics[width=1\textwidth]{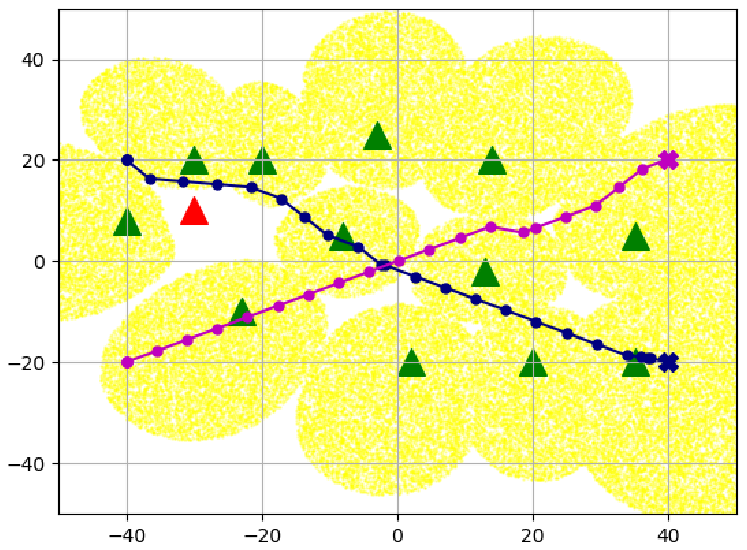}
			\subcaption{\scriptsize Jammer at (-30,10) with $P_J=1$W.  }
		\end{minipage}
		\caption{\small Illustrations of  trajectories in environments, where jammer does not exit or the jammer is located at different positions.
			\label{Fig:diffenv_comparison}  \normalsize}
	\end{figure*}

	Fig. \ref{Fig:diffenv_comparison} displays the illustrations of real-time navigation trajectories, while also depicting the trajectory changes due to the presence of jammers. In the illustrations of the environment and trajectories, the GBSs and the jammer are marked by green triangles and a red triangle, respectively. The yellow areas indicate the communication coverage zones where the agents are able to connect with the cellular network (i.e., $\mS_{r} \geq \mT_{s} $). UAV trajectories are displayed as lines with dots in different colors, and the destinations are marked with crosses. It can be observed from the figure that jammers can generate disconnection/no-coverage zones (the white areas), and different jammer locations lead to different impact. Due to the jammer presence, the UAVs have to make turns and avoid flying through the no-coverage zones, leading to different trajectories.

	To evaluate the performance, we choose the following metrics: 1) success rate, where success indicates one UAV arriving at its destination successfully; 2) disconnection rate, where a disconnection means one UAV being disconnected continuously more than $\mT_t$; 3) collision rate, which quantifies the collisions among UAVs.
	Table \ref{Table:initial_comparison} provides performance comparisons  with benchmarks, considering the above the three metrics. Two benchmarks are chosen: \emph{outdated  map} method in which the UAVs navigate with the trained value network plus the outdated radio map of the environment without jammers; \emph{perfectly-updated map} method in which the UAVs navigate with the value network plus the perfect radio map of the current environment with jammers. The outdated map method does not react to the existence of the jammer, and leads to the performance lower bounds. The perfectly-updated map method is ideal, and achieves the performance upper bound. From the results in the table, we notice that low success rates and high disconnection rates are experienced when the outdated SINR map (which disregards the presence of the jammer) is used. Hence, jammer can have significant impact on the performance. On the other hand, we observe that if the perfect SINR map (which takes into account the interference introduced by the jammer) is utilized, success rates reach above $94\%$ and disconnection rates fall below $3.3\%$. Hence, perfect knowledge of the SINR map is an effective defensive measure against jamming attacks. We note that even with the perfect map, path planning is performed using the deep RL agent.  In our proposed approach, we have the deep RL agent operating with an online SINR learning algorithm. In this case, SINR map can be learned albeit imperfectly. We see in Table I that the proposed approach achieves almost the same performance levels as in the case of the perfectly updated map, and hence leads to effective jamming resiliency in an online fashion. Finally, we remark that in all cases, collision rates are very small, indicating the efficacy of the RL agent operating under collision constraints.
	
	\begin{table*}
		\centering
		\caption{Performance comparison in terms of success rate, disconnection rate, and collision rate. }
		\label{Table:initial_comparison}
		\footnotesize
		\begin{tabular}{|l|l|l|l|l|l|l|l|l|l|}
			\hline
			\multicolumn{1}{|c|}{\multirow{2}{*}{}}                                                      & \multicolumn{3}{c|}{Success Rate (\%)}           & \multicolumn{3}{c|}{Disconnection Rate (\%)}           & \multicolumn{3}{c|}{Collision Rate (\%)}           \\ \cline{2-10}
			\multicolumn{1}{|c|}{}           &
			\begin{tabular}[c]{@{}c@{}}Proposed\\ Algorithm\end{tabular} & \begin{tabular}[c]{@{}c@{}}Outdated\\ Map\end{tabular} & \begin{tabular}[c]{@{}c@{}}Perfectly-\\ Updated Map\end{tabular} & \begin{tabular}[c]{@{}c@{}}Proposed\\ Algorithm\end{tabular} & \begin{tabular}[c]{@{}c@{}}Outdated\\ Map\end{tabular} & \begin{tabular}[c]{@{}c@{}}Perfectly-\\ Updated Map\end{tabular} & \begin{tabular}[c]{@{}c@{}}Proposed\\ Algorithm\end{tabular} & \begin{tabular}[c]{@{}c@{}}Outdated\\ Map\end{tabular} & \begin{tabular}[c]{@{}c@{}}Perfectly-\\ Updated Map\end{tabular} \\ \hline
			
			\begin{tabular}[c]{@{}l@{}}Jammer at (0,0)\\ $P_J=1$W\end{tabular}      &   \multicolumn{1}{c|}{92.3\% }& \multicolumn{1}{c|}{65.3\%}   &  \multicolumn{1}{c|}{94.6\% }     & \multicolumn{1}{c|}{5.6\%}    & \multicolumn{1}{c|}{33.7\%}    & \multicolumn{1}{c|}{3.2\%}       & \multicolumn{1}{c|}{1.8\%}   & \multicolumn{1}{c|}{1.0\%}   & \multicolumn{1}{c|}{1.9\%}   \\ \hline
			\begin{tabular}[c]{@{}l@{}}Jammer at (25,-10)\\ $P_J=1$W\end{tabular}                            & \multicolumn{1}{c|}{94.5\% }   & \multicolumn{1}{c|}{71.1\% }   & \multicolumn{1}{c|}{94.5\%}   & \multicolumn{1}{c|}{4.0\%}   &  \multicolumn{1}{c|}{27.5\%}  & \multicolumn{1}{c|}{3.3\%}   & \multicolumn{1}{c|}{1.5\%}   & \multicolumn{1}{c|}{1.4\%}   &  \multicolumn{1}{c|}{2.2\%}  \\ \hline
			
			\begin{tabular}[c]{@{}l@{}}Jammer at (-30,10)\\ $P_J = 1$W\end{tabular}                          &  \multicolumn{1}{c|}{95.75\%}   & \multicolumn{1}{c|}{78.4\%}   & \multicolumn{1}{c|}{96.6\%}    &    \multicolumn{1}{c|}{3.75\% }  & \multicolumn{1}{c|}{21.2\%}   &  \multicolumn{1}{c|}{2.7\%}   &  \multicolumn{1}{c|}{0.5\%}  & \multicolumn{1}{c|}{0.4\%}   & \multicolumn{1}{c|}{0.6\%}   \\ \hline
			\begin{tabular}[c]{@{}l@{}}Jammer at (0,0)\\ $P_J=0.5$W\end{tabular}                            &  \multicolumn{1}{c|}{92.3\%} & \multicolumn{1}{c|}{73.3\%}   & \multicolumn{1}{c|}{95.8\%}   &  \multicolumn{1}{c|}{6.2\% }    & \multicolumn{1}{c|}{25.5\% }   & \multicolumn{1}{c|}{ 1.9\%}   &  \multicolumn{1}{c|}{1.2\%}   & \multicolumn{1}{c|}{1.2\%}   & \multicolumn{1}{c|}{1.6\% }  \\ \hline
%			\begin{tabular}[c]{@{}l@{}}Jammer at (25,20)\\ $P_J=0.5$W\end{tabular}                            &                         &    &    &                         &    &    &                         &    &    \\ \hline
%			\begin{tabular}[c]{@{}l@{}}Jammer at (-40,-10)\\ $P_J=0.5$W\end{tabular}                            &                         &    &    &                         &    &    &                         &    &    \\ \hline
		\end{tabular}
	\end{table*}

\section{Conclusion}
In this paper, we have addressed jamming-resilient path planning and trajectory designs  for multiple cellular-connected UAVs, while satisfying   wireless connectivity requirements with GBSs and collision avoidance constraints, in the presence of a dynamic jammer. We  formulated the problem as a sequential  decision making problem in discrete time domain and  addressed it via deep RL. Then, we proposed an offline TD learning algorithm for the RL agent with online SINR mapping  to solve the problem. More specifically, a value network has been trained offline by TD method  to encode the interactions among the UAVs and between the UAVs and the environment; and an online SINR mapping DNN has been constructed and trained by supervised learning, to encode the influence of the jammer.
Numerical results have shown that, without any information on the jammer, the proposed algorithm can achieve performance levels close to that of the ideal scenario with the perfect SINR-map.   Hence, real-time navigation for multi-UAVs can be efficiently performed with high success rates, and collisions are avoided.

	\bibliographystyle{IEEEtran}
	\bibliography{compresensive2}

% % % % % % % % Bios % % % % % % % % % %
%	\begin{IEEEbiography}[{\includegraphics[width=1in,height=1.25in,clip,keepaspectratio]{Xueyuan_Wang.eps}}]{Xueyuan Wang}
%		received the B.S. degree in electrical and electronics engineering from Beijing University of Posts and Telecommunications, Beijing, China, in 2013 and the M.S. degree in electrical engineering from Syracuse  University, Syracuse, NY, in 2016. She is currently pursuing  the Ph.D. degree in the Department of Electrical Engineering and Computer Science, Syracuse University. Her primary research interests include millimeter wave communications, heterogeneous cellular networks, stochastic geometry applications, unmanned aerial vehicles, wireless information and power transfer, MIMO, hybrid beamforming and machine learning.
%	\end{IEEEbiography}

\end{document}